\documentclass[11pt,a4paper]{article}
\usepackage[hyperref]{acl2019}
\usepackage{times}
\usepackage{latexsym}
\usepackage[utf8]{inputenc}
\usepackage{amsmath}
\usepackage{amsfonts}
\usepackage{booktabs}
\usepackage{algorithm}
\usepackage[noend]{algpseudocode}
\usepackage{comment}
\usepackage{bbm}
\usepackage{natbib}
\usepackage{stmaryrd}
\usepackage{empheq}
\usepackage{mathtools}
\usepackage{graphicx,multirow}
\usepackage[normalem]{ulem}
\usepackage{subcaption}
\usepackage{siunitx}
\usepackage{enumitem}

\algrenewcommand\algorithmicindent{1.0em}

\newcommand{\E}{\mathbb{E}}

\usepackage{url}

\aclfinalcopy 
\def\aclpaperid{1606} 

\title{Self-Regulated Interactive Sequence-to-Sequence Learning}

\author{Julia Kreutzer \\
  Computational Linguistics \\
  Heidelberg University \\
  Germany\\
  \texttt{\small kreutzer@cl.uni-heidelberg.de} \\\And
  Stefan Riezler \\
 Computational Linguistics \& IWR \\
  Heidelberg University \\
 Germany\\
  \texttt{\small riezler@cl.uni-heidelberg.de} \\}

\date{}

\begin{document}
\maketitle
\begin{abstract}
Not all types of supervision signals are created equal: Different types of feedback have different costs and effects on learning. We show how self-regulation strategies that decide when to ask for which kind of feedback from a teacher (or from oneself) can be cast as a learning-to-learn problem leading to improved cost-aware sequence-to-sequence learning. In experiments on interactive neural machine translation, we find that the self-regulator discovers an $\epsilon$-greedy strategy for the optimal cost-quality trade-off by mixing different feedback types including corrections, error markups, and self-supervision. Furthermore, we demonstrate its robustness under domain shift and identify it as a promising alternative to active learning.
\end{abstract}

\section{Introduction}

The concept of self-regulation has been studied in educational research \cite{HattieTimperley:07,HattieDonoghue:16}, psychology \cite{ZimmermanSchunk:89,Panadero:17}, and psychiatry \cite{Nigg:17}, and was identified as central to successful learning. ``Self-regulated students'' can be characterized as ``becoming like teachers'', in that they have a repertoire of strategies to self-assess and self-manage their learning process, and they know when to seek help and which kind of help to seek. While there is a vast literature on machine learning approaches to meta-learning \cite{SchmidhuberETAL:96}, learning-to-learn \cite{ThrunPratt:98}, or never-ending learning \cite{MitchellETAL:15}, the aspect of learning when to ask for which kind of feedback has so far been neglected in this field.

\begin{figure}
    \centering
    \includegraphics[width=0.8\columnwidth]{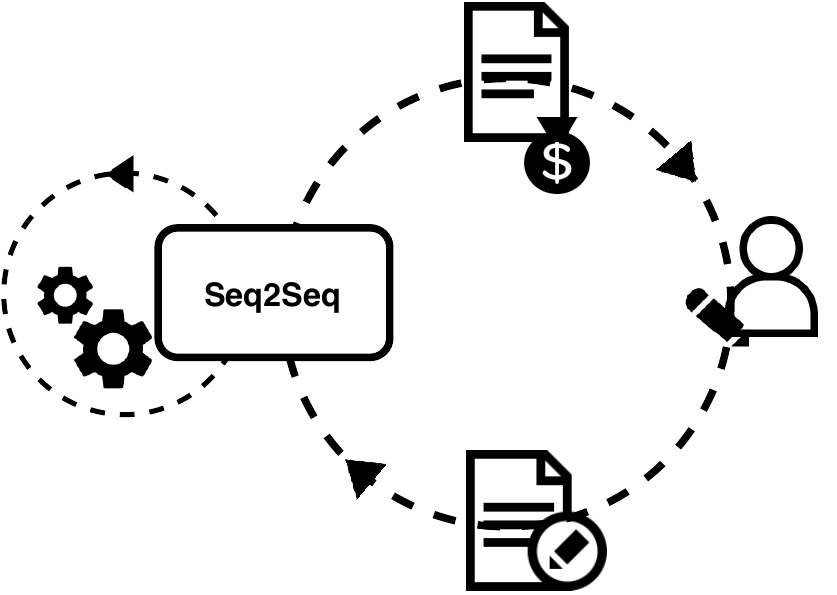}
    \caption{Human-in-the-loop self-regulated learning.}
    \label{fig:arl}
\end{figure}

We propose a machine learning algorithm that uses self-regulation in order to balance the cost and effect of learning 
from different types of feedback. This is particularly relevant for human-in-the-loop machine learning, where human supervision is costly.
The self-regulation module automatically learns which kind of feedback to apply when in training---full supervision by teacher demonstration or correction, weak supervision in the form of positive or negative rewards for student predictions, or a self-supervision signal generated by the student. Figure~\ref{fig:arl} illustrates this learning scenario. The learner, in our case a sequence-to-sequence (Seq2Seq) learner, aims to solve a certain task with the help of a human teacher. For every input it receives for training, it can ask the teacher for feedback to its own output, or supervise itself by training on its own output, or skip learning on the input example altogether. The self-regulator's policy for choosing feedback types is guided by their cost and by the performance gain achieved by learning from a particular type of feedback.

We apply the self-regulation algorithm to interactive machine translation where a neural machine translation (NMT) system functions as a student which receives feedback simulated from a human reference translation or supervises itself. The intended real-world application is a machine translation personalization scenario where the goal of the human translator is to teach the NMT system to adapt to in-domain data with the best trade-off between feedback cost and performance gain. It can be transferred to other sequence-to-sequence learning tasks such as personalization of conversational AI systems for question-answering or geographical navigation.

Our analysis of different configurations of self-regulation yields the following insights: Perhaps unsurprisingly, the self-regulator learns to balance all types of feedback instead of relying only on the strongest or cheapest option. This is an advantage over active learning strategies that only consider the choice between no supervision and full supervision. Interestingly, though, we find that the self-regulator learns to trade off exploration and exploitation similar to a context-free $\epsilon$-greedy strategy that optimizes $\epsilon$ for fastest learning progress.
Lastly, we show that the learned regulator is robust in a cold-start transfer to new domains, and even shows improvements over fully supervised learning on domains such as literary books where reference translations provide less effective learning signals.

\section{Related Work}

The incorporation of a query's cost into reinforcement learning has been addressed, for example, in the framework of active reinforcement learning \cite{KruegerETAL:16}. The central question in active reinforcement learning is to quantify the long-term value of reward information, however, assuming a fixed cost for each action and every round. Our framework is considerably more complicated by the changing costs for each feedback type on each round. 

A similar motivation for the need of changing feedback in reinforcement learning with human feedback is given in \citet{MacGlashanETAL:17}. The goal of that work is to operationalize feedback schemes such as diminishing returns, differential feedback, or policy shaping. Human reinforcement learning with corrective feedback that can decrease or increase the action magnitude has been introduced in \citet{Celemin2019}. However, none of these works are concerned with the costs that are incurred when eliciting rewards from humans, nor do they consider multiple feedback modes.

Our work is connected to active learning, for example, to approaches that use reinforcement learning to learn a policy for a dynamic active learning strategy \cite{FangETAL:17}, or to learn a curriculum to order noisy examples \cite{KumarETAL:19}, or to the approach of \citet{liu-buntine-haffari:2018:K18-1} who use imitation learning to select batches of data to be labeled. However, the action space these approaches consider is restricted to the decision whether or not to select particular data and is designed for a fixed budget, neither do they incorporate feedback cost in their frameworks. 
As we will show, our self-regulation strategy outperforms active learning based on uncertainty sampling \cite{SettlesCraven:08,PerisCasacuberta:18} and our reinforcement learner is rewarded in such a way that it will produce the best system as early as possible. 

Research that addresses the choice and the combination of different types of feedback is situated in the area between reinforcement and imitation learning \cite{RanzatoETAL:16,ChengETAL:18}. Instead of learning how to mix different supervision signals, these approaches assume fixed schedules.

Further connections between our work on learning with multiple feedback types can be drawn to various extensions of reinforcement learning by multiple tasks \cite{JaderbergETAL:17}, 
multiple loss functions \cite{WuETAL:18}, or multiple policies \cite{SmithETAL:18}.

Feedback in the form of corrections \cite{turchi2017continuous}, error markings \cite{Domingo2017}, or translation quality judgments \cite{LamETAL:18} has been successfully integrated in simulation experiments into interactive-predictive machine translation. Again, these works do not consider automatic learning of a policy for the optimal choice of feedback.

\section{Self-Regulated Interactive Learning}

In this work, we focus on the aspect of self-regulated learning that concerns the ability to decide which type of feedback to query from a teacher (or oneself) for most efficient learning depending on the context. 
In our human-in-the-loop machine learning formulation, we focus on two contextual aspects that can be measured precisely: quality and cost. 
The self-regulation task is to optimally balance human effort and output quality.

We model self-regulation as an active reinforcement learning problem with dynamic costs, where in each state, i.e. upon receiving an input, the regulator has to choose an action, here a feedback type, and pay a cost. The learner receives feedback of that type from the human to improve its prediction. Based on the effect of this learning update, the regulator's actions are reinforced or penalized, so that it improves its choice for future inputs. 

In the following, we first compare training objectives for a Seq2Seq learner from various types of feedback (\S\ref{sub:seq2seq}), then introduce the self-regulator module (\S\ref{sub:regulator}), and finally combine both in the self-regulation algorithm (\S\ref{sub:algo}).

\subsection{Seq2Seq Learning}\label{sub:seq2seq}
Let $x = x_1 \dots x_S$ be a sequence of indices over a source vocabulary $\mathcal{V}_{\textsc{Src}}$, and $y = y_1 \dots y_T$ a sequence of indices over a target vocabulary $\mathcal{V}_{\textsc{Trg}}$. The goal of sequence-to-sequence learning is to learn a function for mapping an input sequence $x$  into an output sequences $y$. Specifically, for the example of machine translation, where $y$ is a translation of $x$, the model, parametrized by a set of weights $\theta$, learns to maximize $p_{\theta}(y \mid x)$. This quantity is further 
factorized into conditional probabilities over single tokens:
\begin{align*}
    p_{\theta}(y \mid x) = \prod^{T}_{t=1}{p_{\theta}(y_t \mid x; y_{<t})}.
\end{align*}
The distribution $p_{\theta}(y_t \mid x; y_{<t})$ is defined by the neural model's softmax-normalized output vector
\begin{align*}
    p_{\theta}(y_t \mid x; y_{<t}) = \texttt{softmax}(\text{NN}_{\theta}(x; y_{<t})).
\end{align*}
There are various options for building the architecture of the neural model $\text{NN}_{\theta}$, such as recurrent \cite{sutskever2014sequence}, convolutional \cite{gehring2017convolutional} or attentional \cite{vaswani2017attention} encoder-decoder architectures (or a mix thereof \cite{bestofbothworlds}). Regardless of their architecture, there are multiple ways of interactive learning that can be applied to neural Seq2Seq learners.

\paragraph{Learning from Corrections (\textsc{Full}).}
Under full supervision, i.e., when the learner receives a fully corrected output $y^*$ for an input $x$, cross-entropy minimization (equivalent to maximizing the likelihood of the data $\mathcal{D}$ under the current model) considers the following objective:
\begin{align*}
    J^{\textsc{Full}}(\theta) = \frac{1}{|\mathcal{D}|} \sum_{(x,y^*) \in \mathcal{D}}  - \log p_{\theta}(y^* \mid x) .
\end{align*}
The stochastic gradient of this objective is
\begin{align*}
g^{\textsc{Full}}_{\theta}(x, y^*) & = -  \nabla_{\theta} \log p_{\theta}(y^* \mid x) , 
\end{align*}
constituting an unbiased estimate of the gradient
\begin{align*}
    \nabla_{\theta} J^{\textsc{Full}} =& \E_{(x,y^*) \sim \mathcal{D}}  \left[g^{\textsc{Full}}_{\theta}(x, y^*)\right].
\end{align*}
A local minimum can be found by performing stochastic gradient descent on $g^{\textsc{Full}}_{\theta}(x, y^*)$.
This training objective is the standard in supervised learning when training with human-generated references or for online adaptation to post-edits \cite{turchi2017continuous}.

\paragraph{Learning from Error Markings (\textsc{Weak}).} 
\citet{petrushkov-khadivi-matusov:2018:Short} presented chunk-based binary feedback as a low-cost alternative to full corrections.
In this scenario the human teacher marks the correct parts of the machine-generated output $\hat{y}$. As a consequence every token in the output receives a reward $\delta_t$, either $\delta_t=1$ if marked as correct, or $\delta_t=0$ otherwise. The objective of the learner is to maximize the likelihood of the correct parts of the output, or equivalently, to minimize
\begin{equation*}
\resizebox{\hsize}{!}{
$\textstyle J^{\textsc{Weak}}(\theta) = \frac{1}{|\mathcal{D}|} \sum_{(x,\hat{y}) \in \mathcal{D}} \sum_{t=1}^{T}{- \delta_t \, \log p_{\theta}(\hat{y}_t \mid x; \hat{y}_{<t})}$
}
\end{equation*}
where the stochastic gradient is
\begin{align*}
\textstyle 
g^{\textsc{Weak}}_{\theta}(x, \hat{y}) &= - \sum_{t=1}^{T}\delta_t \cdot \nabla_{\theta} \log p_{\theta}(\hat{y}_t \mid x; y_{<t}) \\
\nabla_{\theta} J^{\textsc{Weak}} &= \E_{(x,\hat{y}) \sim \mathcal{D}}  \left[g^{\textsc{Weak}}_{\theta}(x, \hat{y})\right].
\end{align*}
The tokens $\hat{y}_t$ that receive $\delta_t=1$ are part of the correct output $y^{*}$, so the model receives a hint how a corrected output should look like. Although the likelihood of the incorrect parts of the sequence does not weigh into the sum, they are contained in the context of the correct parts (in $y_{<t}$).
\paragraph{Self-Supervision (\textsc{Self}).}
Instead of querying the teacher for feedback, the learner can also choose to learn from its own output, that is, to learn from self-supervision. 
The simplest option is to treat the learner's output as if it was correct, but that quickly leads to overconfidence and degeneration. \citet{ClarkETAL:2018} proposed a cross-view training method: the learner's original prediction is used as a target for a weaker model that shares parameters with the original model.
We adopt this strategy by first producing a target sequence $\hat{y}$ with beam search and then weaken the decoder through attention dropout with probability $p_{att}$. 
The objective is to minimize the negative likelihood of the original target under the weakened model
\begin{align*}
\textstyle
        J^{\textsc{Self}}(\theta) &= \frac{1}{|\mathcal{D}|} \sum_{(x,\hat{y}) \in \mathcal{D}}  - \log p^{p_{att}}_{\theta}(\hat{y} \mid x),
\end{align*}
where the stochastic gradient is
\begin{align*}
\textstyle
    g^{\textsc{Self}}_{\theta}(x, \hat{y}) &= - \nabla_{\theta} \log p^{p_{att}}_{\theta}(\hat{y} \mid x) \\
       \nabla_{\theta} J^{\textsc{Self}}  &= \E_{(x,\hat{y}) \sim \mathcal{D}} \left[  g^{\textsc{Self}}_{\theta}(x, \hat{y}) \right].
\end{align*}

\paragraph{Combination.}
For self-regulated learning, we also consider a fourth option ($\textsc{None}$): the option to ignore the current input. Figure~\ref{fig:grad-general} summarizes the stochastic gradients for all cases. 

\begin{figure}[h]
\begin{equation*}
    \boxed{
    \begin{aligned}
 g^s_{\theta}(x, y) &= - \sum_{t=1}^{T} f_t  \cdot \nabla_{\theta} \log p^{drop}_{\theta}(y_t \mid x_t; y_{<t}),\\
   \text{with } y &= \begin{cases}
        y^* & \text{if } s=\textsc{Full} \\
        \hat{y} & \text{otherwise}, \\
        \end{cases}\\
    drop &= \begin{cases}
        p_{\text{att}} & \text{if }s=\textsc{Self} \\
        0 & \text{otherwise,} \\
        \end{cases}\\
    \text{and }
    f_t &= \begin{cases}
        1 & \text{if } s\in\{\textsc{Full},\textsc{Self}\} \\
        \delta_t & \text{if }s=\textsc{Weak} \\
        0 & \text{if }s=\textsc{None} \\ 
        \end{cases}
    \end{aligned}
    }
\end{equation*}
\caption{Stochastic gradients for the Seq2Seq learner in dependence of feedback type $s$.}
\label{fig:grad-general}
\end{figure}
In practice, Seq2Seq learning shows greater stability for mini-batch updates than online updates on single training samples. Mini-batch self-regulated learning can be achieved by accumulating stochastic gradients for a mini-batch of size $\mathcal{B}$ before updating $\theta$ with an average of these stochastic gradients, which we denote as $g^{{s}_{[1:\mathcal{B}]}}_{\theta}(x_{[1:\mathcal{B}]}, y_{[1:\mathcal{B}]}) = \frac{1}{\mathcal{B}} \sum^{\mathcal{B}}_{i=1} g_{\theta}^{s_i}(x_i, y_i)$.

\subsection{Learning to Self-Regulate}\label{sub:regulator}
The regulator is another neural model $q_{\phi}$ that is optimized for the quality-cost trade-off of the Seq2Seq learner. Given an input $x_i$ and the Seq2Seq's hypothesis $\hat{y}_i$, it chooses an action, here a supervision mode $s_i \sim q_{\phi}(s \mid x_i, \hat{y}_i)$. This choice of feedback determines the update of the Seq2Seq learner (Figure~\ref{fig:grad-general}).
The regulator is rewarded by the ratio between the cost $c_i$ of obtaining the feedback $s_i$ and the quality improvement $\Delta(\theta_i, \theta_{i-1})$ caused by updating the Seq2Seq learner with the feedback:
\begin{align} \label{eq:reward}
    r(s_i,  x_i, \theta_i) = \frac{\Delta(\theta_i, \theta_{i-1})}{c_i+\alpha}.
\end{align}
$\Delta(\theta_i, \theta_{i-1})$ is measured as the difference in validation score achieved before and after the learner's update \cite{FangETAL:17}, and $c_i$ as the cost of user edits. Adding a small constant cost $\alpha$ to the actual feedback cost ensures numerical stability. This meta-parameter can be interpreted as representing a basic cost for model updates of any kind. 

The objective for the regulator is to maximize the expected reward defined in Eq.~\ref{eq:reward}: 
\begin{align*} 
    J^{\textsc{Meta}}(\phi) &= \E_{x \sim p(x), s \sim q_{\phi}(s \mid x, \hat{y})} \left[ r(s,  x, \theta) \right]. 
\end{align*}

The full gradient of this objective is estimated by the stochastic gradient for sampled actions \cite{Williams:92}:
\begin{align}\label{eq:metagrad}
  \textstyle 
  g^{\textsc{Meta}}_{\phi}(x, \hat{y}, s) &= r \cdot \nabla_{\phi} \log q_{\phi}(s \mid x, \hat{y}).
\end{align}

Note that the reward contains the immediate improvement after one update of the Seq2Seq learner and not the overall performance in hindsight. This is an important distinction to classic expected reward objectives in reinforcement learning since it biases the regulator towards actions that have an immediate effect, which is desirable in the case of interaction with a human. 
However, since Seq2Seq learning requires updates and evaluations based on mini-batches, the regulator update also needs to be based on mini-batches of predictions, leading to the following specification of Eq.~\eqref{eq:metagrad} for a mini-batch $j$:
\begin{align}\label{eq:metagrad-batch}
    g&^{\textsc{Meta}}_{\phi}(x_{[1:\mathcal{B}]}, \hat{y}_{[1:\mathcal{B}]}, s_{[1:\mathcal{B}]}) \\ \nonumber
    &= \frac{1}{\mathcal{B}} \sum_{i=1}^{\mathcal{B}} g^{\textsc{Meta}}_{\phi}(x_i, \hat{y}_i, s_i) \\ \nonumber
    &= \Delta(\theta_j, \theta_{j-1}) \frac{1}{\mathcal{B}} \sum_{i=1}^{\mathcal{B}}{\frac{ \nabla_{\phi} \log q_{\phi}(s_i \mid x_i, \hat{y}_i)}{c_i+\alpha}}.
\end{align}
While mini-batch updates are required for stable Seq2Seq learning, they hinder the regulator from assigning credit for model improvement to individual elements within the mini-batch. 

\subsection{Algorithm}\label{sub:algo}
Algorithm~\ref{alg:sr} presents the proposed online learning algorithm with model updates cumulated over mini-batches. On arrival of a new input, the regulator predicts a feedback type 
in line~\ref{line:regulator}. According to this prediction, the environment/user is asked for feedback for the Seq2Seq's prediction at cost $c_i$ (line~\ref{line:feedback}).
The Seq2Seq model is updated on the basis of the feedback and mini-batch of stochastic gradients computed as summarized in Figure~\ref{fig:grad-general}.
In order to reinforce the regulator, the Seq2Seq model's improvement (line~\ref{line:eval}) is assessed, and the parameters of the regulator are updated (line~\ref{line:reg-update}, Eq.~\ref{eq:metagrad-batch}).
Training ends when the data stream or the provision of feedback ends. The intermediate Seq2Seq evaluations can be re-used for model selection (early stopping). In practice, these evaluations can either be performed by validation on a held-out set (as in the simulation experiments below) or by human assessment.

\begin{algorithm}[t]
\begin{algorithmic}[1]
  	\Require{Initial Seq2Seq $\theta_0$, regulator $\phi_0$, $\mathcal{B}$} 
  \State{$j \gets 0$} 
  \While{inputs and human available}
    \State{$j \gets j+1$}
    \For{$i \gets 1 \text{ to } \mathcal{B}$} 
  	    \State{Observe input $x_i$, Seq2Seq output $\hat{y}_i$} 
  	    \State{Choose feedback: $s_i \sim q_{\phi}(s \mid x_i, \hat{y}_i)$} \label{line:regulator}
  	    \State{Obtain feedback $f_i$ of type $s_i$ at cost $c_i$} \label{line:feedback}
  	\EndFor
  	\State{Update $\theta$ with $g^{s_{[1:\mathcal{B}]}}_{\theta}(x_{[1:\mathcal{B}]}, \hat{y}_{[1:\mathcal{B}]})$ }  \label{line:gradient} 
  	\State{Measure improvement $\Delta(\theta_{j}, \theta_{j-1})$} \label{line:eval}  
  	\State{Update $\phi$ with $g^{\textsc{Meta}}_{\phi} (x_{[1:\mathcal{B}]}, \hat{y}_{[1:\mathcal{B}]}, s_{[1:\mathcal{B}]})$ } \label{line:reg-update} 
  	 \EndWhile

\end{algorithmic}
  \caption{{Self-Regulated Interactive Seq2Seq}}
  \label{alg:sr}
\end{algorithm}

\paragraph{Practical Considerations.}
The algorithm does not introduce any additional hyperparameters beyond standard learning rates, architecture design and mini-batch sizes that have to be tuned.
As proposed in \citet{petrushkov-khadivi-matusov:2018:Short} or \citet{ClarkETAL:2018}, targets $\hat{y}$ are pre-generated offline with the initial $\theta_0$, which we found crucial for the stability of the learning process.
The evaluation step after the Seq2Seq update is an overhead that comes with meta-learning, incurring costs depending on the decoding algorithm and the evaluation strategy. 
However, Seq2Seq updates can be performed in mini-batches, and the improvement is assessed after a mini-batch of updates, as discussed above.

\section{Experiments}\label{sec:experiments}
The main research questions to be answered in our experiments are:
\begin{enumerate}
    \item Which strategies does the regulator develop?
     \item How well does a trained regulator transfer across domains?
    \item How do these strategies compare against (active) learning from a single feedback type?
\end{enumerate}
We perform experiments for interactive NMT, where a general-domain NMT model is adapted to a specific domain by learning from the feedback of a human translator. 
This is a realistic interactive learning scenario where cost-free pre-training on a general domain data is possible, but each feedback generated by the human translator in the personalization step incurs a specific cost.
In our experiment, we use human-generated reference translations to simulate both the cost of human feedback and to measure the performance gain achieved by model updates. 

\subsection{Experimental Setup}
\paragraph{Seq2Seq Architecture.}
Both the Seq2Seq learner and the regulator are based on LSTMs \cite{hochreiter1997long}. 
The Seq2Seq has four bi-directional encoder and four decoder layers with 1024 units each, embedding layers of size 512. It uses \citet{luong-pham-manning:2015:EMNLP}'s input feeding and output layer, and global attention with a single feed forward layer \cite{BahdanauETAL:15}.

\paragraph{Regulator Architecture.}
The regulator consists of LSTMs on two levels: Inspired by Siamese Networks \cite{bromley1994signature}, a bi-directional LSTM encoder of size 512 separately reads in both the current input sequence and the beam search hypothesis generated by the Seq2Seq. The last state of encoded source and hypothesis sequence and the previous output distribution are concatenated to form the input to a higher-level regulator LSTM of size 256. This LSTM updates its internal state and predicts a score for every feedback type for every input in the mini-batch. The feedback for each input is chosen by sampling from the distribution obtained by softmax normalization of these scores.
The embeddings of the regulator are initialized by the Seq2Seq's source embeddings and further tuned during training.
The model is implemented in the JoeyNMT\footnote{\url{https://github.com/joeynmt/joeynmt}} framework based on PyTorch \citep{JoeyNMT}.\footnote{Code: \url{https://github.com/juliakreutzer/joeynmt/tree/acl19}}

\paragraph{Data.}
We use three parallel corpora for German-to-English translation: a general-domain data set from the WMT2017 translation shared task for Seq2Seq pre-training, TED talks from the IWSLT2017 evaluation campaign for training the regulator with simulated feedback, and the Books corpus from the OPUS collection \cite{TIEDEMANN12.463} for testing the regulator on another domain. Data pre-processing details and splits are given in \S\ref{app:data}. The joint vocabulary for Seq2Seq and the regulator consists of 32k BPE sub-words \cite{BPE} trained on WMT.

\paragraph{Training.}
The Seq2Seq model is first trained on WMT with Adam \cite{KingmaBa:14} on mini-batches of size 64, an initial learning rate \num{1e-4} that is halved when the loss does not decrease for three validation rounds.
Training ends when the validation score does not increase any further (scoring 29.08 BLEU on the WMT test). This model is then adapted to IWSLT with self-regulated training for one epoch, with online human feedback simulated from reference translations. The mini-batch size is reduced to 32 for self-regulated training to reduce the credit assignment problem for the regulator. The constant cost $\alpha$ (Eq.~\ref{eq:reward}) is set to 1.\footnote{Values $\neq1$ distort the rewards for self-training too much.} When multiple runs are reported, the same set of random seeds is used for all models to control the order of the input data. The best run is evaluated on the Books domain for testing the generalization of the regulation strategies.

\paragraph{Simulation of Cost and Performance.}
In our experiments, human feedback and its cost, and the performance gain achieved by model updates, is simulated by using human reference translations.
Inspired by the keystroke  mouse-action ratio (KSMR) \cite{barrachinaETAL:09}, a common metric for measuring human effort in interactive machine translation, we define feedback cost as the sum of costs incurred by character edits and clicks, similar to \citet{PerisCasacuberta:18}.
The cost of a full correction (\textsc{Full}) is the number of character edits between model output and reference, simulating the cost of a human typing.\footnote{As computed by the Python library $\texttt{difflib}$.}
Error markings (\textsc{Weak}) are simulated by comparing the hypothesis to the reference and marking the longest common sub-strings as correct, as proposed by \citet{petrushkov-khadivi-matusov:2018:Short}. As an extension to \citet{petrushkov-khadivi-matusov:2018:Short} we mark multiple common sub-strings as correct if all of them have the longest length. 
The cost is defined as the number of marked words, assuming an interface that allows markings by clicking on words.
For self-training (\textsc{Self}) and skipping training instances we naively assume zero cost, thus limiting the measurement of cost to the effort of the human teacher, and neglecting the effort on the learner's side.
Table~\ref{tab:sim_examples} illustrates the costs per feedback type on a randomly selected set of examples.

\begin{table*}[t]
    \centering
    \resizebox{\textwidth}{!}{
    \begin{tabular}{l|l|ll}
        \toprule
          \multirow{3}{*}{\rotatebox[origin=c]{90}{\textsc{Self}}} & \multirow{3}{*}{0}& $x$ & Sie greift in ihre Geldbörse und gibt ihm einen Zwanziger . \\
         & &$\hat{y}$ & It attacks their wallets and gives him a twist . \\
       & &$y^*$  & She reaches into her purse and hands him a 20 . \\
       \midrule
         \multirow{3}{*}{\rotatebox[origin=c]{90}{\textsc{Weak}}} & \multirow{3}{*}{9} & $x$ & Und als ihr Vater sie sah und sah , wer sie geworden ist , in ihrem vollen Mädchen-Sein , schlang er seine Arme um sie und brach in Tränen aus .  \\
      &  & $\hat{y}$  & And when \underline{her father saw} them \underline{and saw who} became them , in their full girl 's , he swallowed \underline{his arms around} them and broke out in tears .\\
      &  & $y^*$ &  When her father saw her and saw who she had become , in her full girl self , he threw his arms around her and broke down crying .\\
        \midrule
       \multirow{3}{*}{\rotatebox[origin=c]{90}{\textsc{Full}}} & \multirow{3}{*}{59} &$x$ & Und durch diese zwei Eigenschaften war es mir möglich , die Bilder zu erschaffen , die Sie jetzt sehen .\\
      &  & $\hat{y}$ &  And th\sout{rough} th\sout{e}se two \sout{features ,} I was able to create the images you now \sout{see} . \\
      &  & $y^*$  & And \underline{it was wi}th th\underline{o}se two \underline{properties that} I was able to create the images that you \underline{'re seeing right} now . \\
        \bottomrule
    \end{tabular}
    }
    \caption{Examples from the IWSLT17 training set, cost (2nd column) and feedback decisions made by \emph{Reg3}. For weak feedback, marked parts are underlined, for full feedback, the corrections are marked by underlining the parts of the reference that got inserted and the parts of the hypothesis that got deleted.}
    \label{tab:sim_examples}
\end{table*}

We measure the model improvement by evaluating the held-out set translation quality of the learned model at various time steps with corpus BLEU (cased \texttt{SacreBLEU} \cite{sacrebleu}) and measure the accumulated costs. The best model is considered the one that delivers the highest quality at the lowest cost. This trade-off is important to bear in mind since it differs from the standard evaluation of machine translation models, where the overall best-scoring model, regardless of the supervision cost, is considered best. Finally, we evaluate the strategy learned by the regulator on an unseen domain, where the regulator decides which type of feedback the learner gets, but is not updated itself.

\subsection{Results}\label{sec:results}

We compare learning from one type of feedback in isolation against regulators with the following set of actions:
\begin{enumerate} 
    \item \emph{Reg2}: \textsc{Full}, \textsc{Weak}
    \item \emph{Reg3}: \textsc{Full}, \textsc{Weak}, \textsc{Self}
    \item \emph{Reg4}: \textsc{Full}, \textsc{Weak}, \textsc{Self}, \textsc{None}
\end{enumerate}

\begin{figure}[t]
     \centering
 \includegraphics[width=\columnwidth]{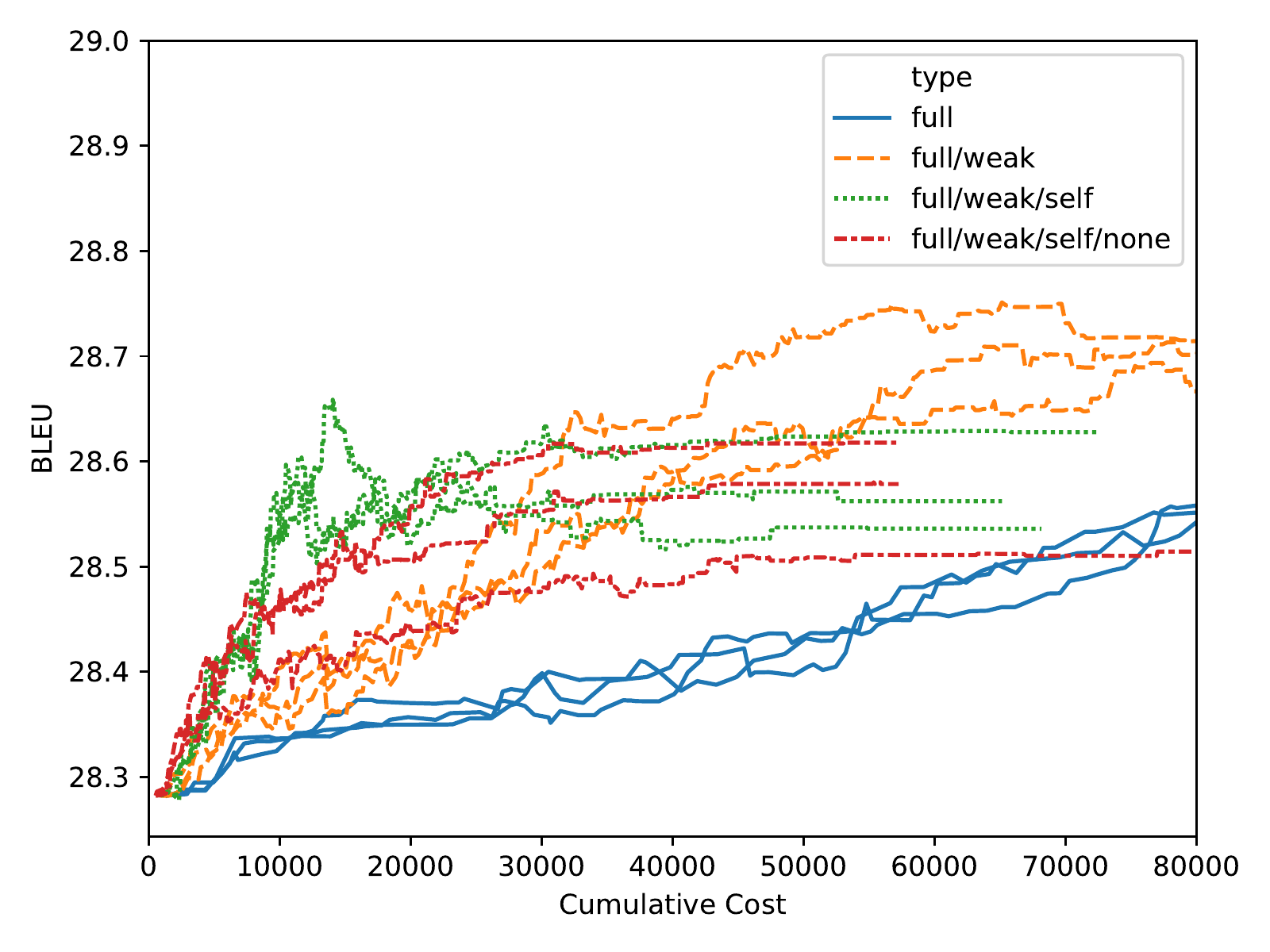}
    \caption{BLEU of regulation variants over cumulative costs. BLEU is computed on the tokenized IWSLT validation set with greedy decoding.}
      \label{fig:bleu-cost}
\end{figure}

\paragraph{Cost vs. Quality.}
Figure~\ref{fig:bleu-cost} compares the improvement in corpus BLEU \cite{papineni2002bleu} 
(corresponding to results in Translation Error Rate (TER, computed by \texttt{pyTER}) \cite{snover2006study})
of regulation variants and full feedback over cumulative costs of up to 80k character edits. Using only full feedback (blue) as in standard supervised learning or learning from post-edits, the overall highest improvement can be reached (visible only after the cutoff of 80k edits; see Appendix \ref{app:iwslt} for the comparison over a wider window of time). However, it comes at a very high cost (417k characters in total to reach +0.6 BLEU). The regulated variants offer a much cheaper improvement, at least until a cumulative cost between 80k (\emph{Reg4}) and 120k (\emph{Reg2}), depending on the feedback options available. The regulators do not reach the quality of the full model since their choice of feedback is oriented towards costs and immediate improvements. By finding a trade-off between feedback types for immediate improvements, the regulators sacrifice long-term improvement. Comparing regulators, \emph{Reg2} (orange) reaches the overall highest improvement over the baseline model, but until the cumulative cost of around 35k character edits, \emph{Reg3} (green) offers faster improvement at a lower cost since it has an additional, cheaper feedback option. Adding the option to skip examples (\emph{Reg4}, red) does not give a benefit. Appendix~\ref{app:offline} lists detailed results for offline evaluation on the trained Seq2Seq models on the IWSLT test set: 
Self-regulating models achieve improvements of 0.4-0.5 BLEU with costs reduced up to a factor of 23 in comparison to the full feedback model.
The reduction in cost is enabled by the use of cheaper feedback, here markings and self-supervision, which in isolation are very successful as well. Self-supervision works surprisingly well and can be recommended for cheap but effective unsupervised domain adaptation for sequence-to-sequence learning.

\paragraph{Self-Regulation Strategies.}

\begin{figure}[h]
    \centering
    \includegraphics[width=\columnwidth]{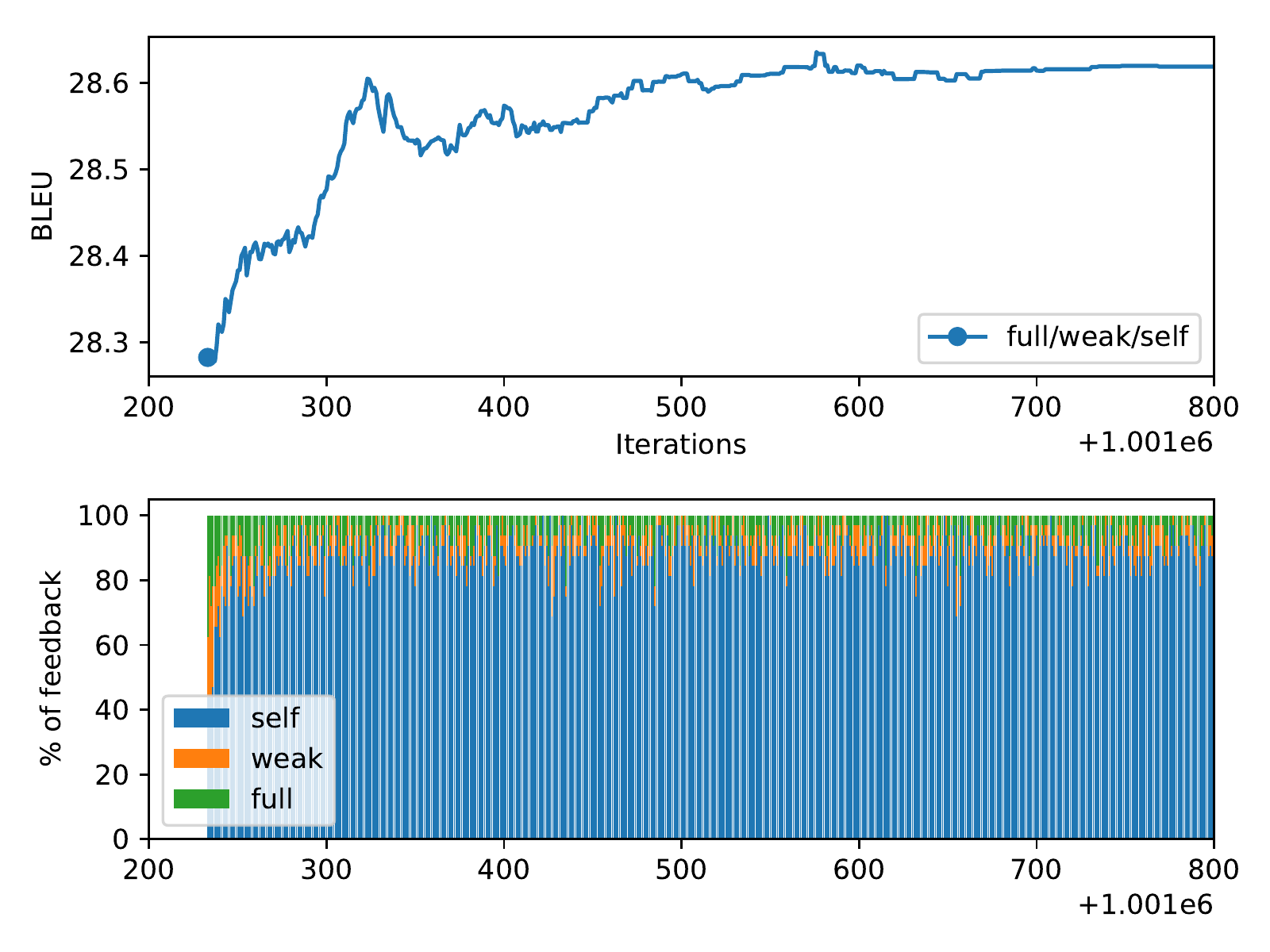}
    \caption{\emph{Reg3} actions as chosen over time, depicted for each iteration. Counting of iterations starts at the previous iteration count of the baseline model. }
    \label{fig:reg_actions}
\end{figure}

Figure~\ref{fig:reg_actions} shows which actions \emph{Reg3} chooses over time when trained on IWSLT. Most often it chooses to do self-training on the current input.
The choice of feedback within one batch varies only slightly during training, with the exception of an initial exploration phase within the first 100 iterations.
In general, we observe that all regulators are highly sensitive to balancing cost and performance, and mostly prefer the cheapest option (e.g., \emph{Reg4} by choosing mostly \textsc{None}) since they are penalized heavily for choosing (or exploring) expensive options (see Eq.~\ref{eq:reward}).
\begin{figure}
    \centering
     \includegraphics[width=\columnwidth]{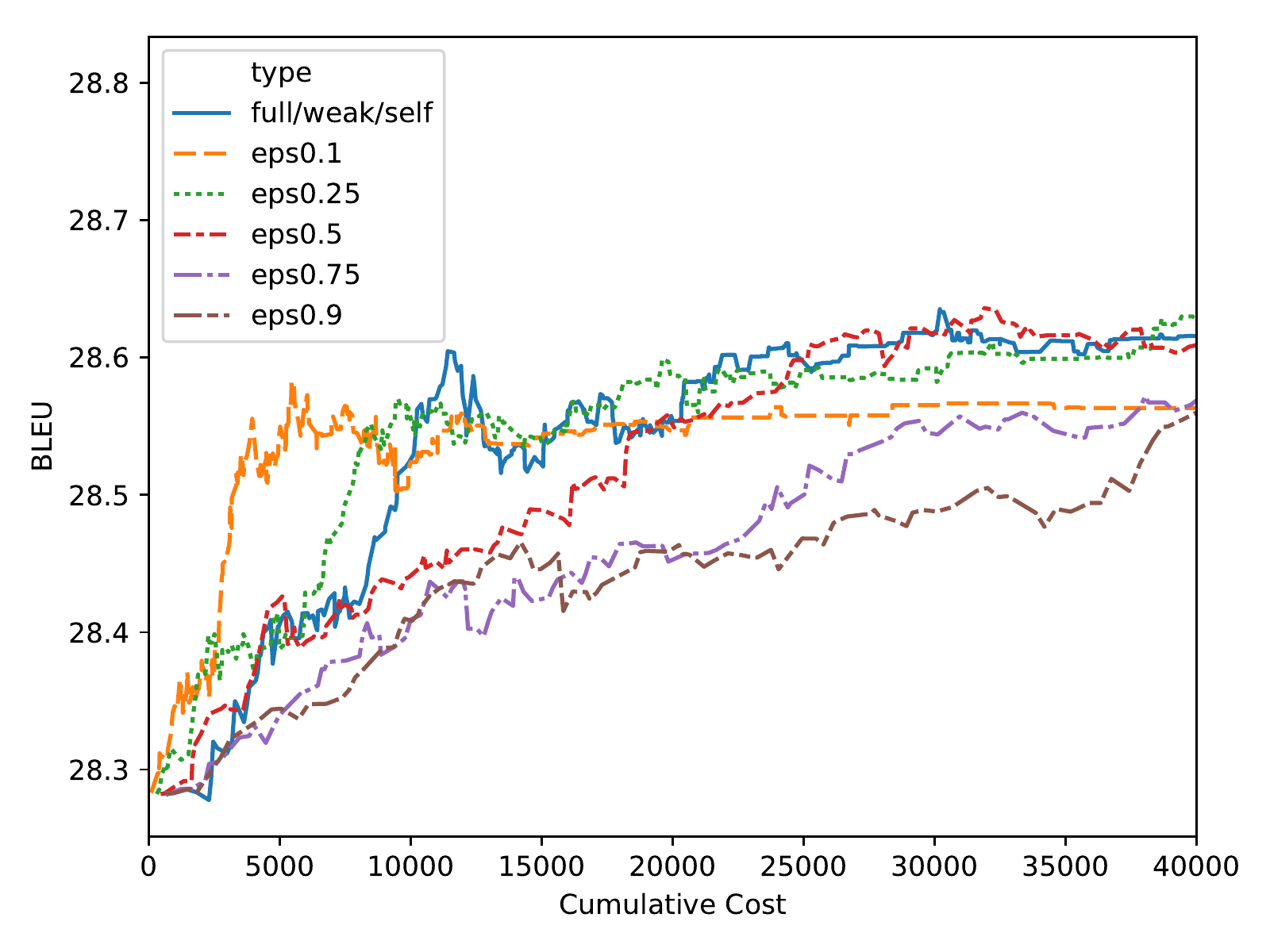}
    \caption{BLEU and cumulative costs on IWSLT for \emph{Reg3} and $\epsilon$-greedy with {$\epsilon\in [0.1, 0.25, 0.5, 0.75, 0.9]$}.} 
    \label{fig:epsilon}
\end{figure}

A further research question is whether and how the self-regulation module takes the input or output context into account.
We therefore compare its decisions to a context-free $\epsilon$-greedy strategy.
The $\epsilon$-greedy algorithm is a successful algorithm for multi-armed bandits \cite{watkins1989learning}. In our case, the arms are the four feedback types. They are chosen based on their reward statistics, here the average empirical reward per feedback type $Q_i(s) = \frac{1}{N_i(s)} \sum_{0, \dots, i}{r(s_i)}$. With probability $1-\epsilon$, the algorithm selects the feedback type with the highest empirical reward (exploitation), otherwise picks one of the remaining arms at random (exploration).
In contrast to the neural regulator model, $\epsilon$-greedy decides solely on the basis of the reward statistics and has no internal contextual state representation. 
The comparison of \emph{Reg3} with $\epsilon$-greedy for a range of values for $\epsilon$ in Figure~\ref{fig:epsilon} shows that learned regulator behaves indeed very similar to an $\epsilon$-greedy strategy with $\epsilon=0.25$. $\epsilon$-greedy variants with higher amounts of exploration show a slower increase in BLEU, while those with more exploitation show an initial steep increase that flattens out, leading to overall lower BLEU scores. The regulator has hence found the best trade-off, which is an advantage over the $\epsilon$-greedy algorithm where the $\epsilon$ hyperparameter requires dedicated tuning.
Considering the $\epsilon$-greedy-like strategy of the regulator and the strong role of the cost factor shown in Figure~\ref{fig:reg_actions}, the regulator module does not appear to choose individual actions based e.g., on the difficulty of inputs, but rather composes mini-batches with a feedback ratio according to the feedback type's statistics.
This confirms the observations of \citet{PerisCasacuberta:18}, who find that the subset of instances selected for labeling is secondary---it is rather the mixing ratio of feedback types that matters. This finding is also consistent with the mini-batch update regime that forces the regulator to take a higher-level perspective and optimize the expected improvement at the granularity of (mini-batch) updates rather than at the input level.

\paragraph{Domain Transfer.}
\begin{figure}
    \centering
    \includegraphics[width=\columnwidth]{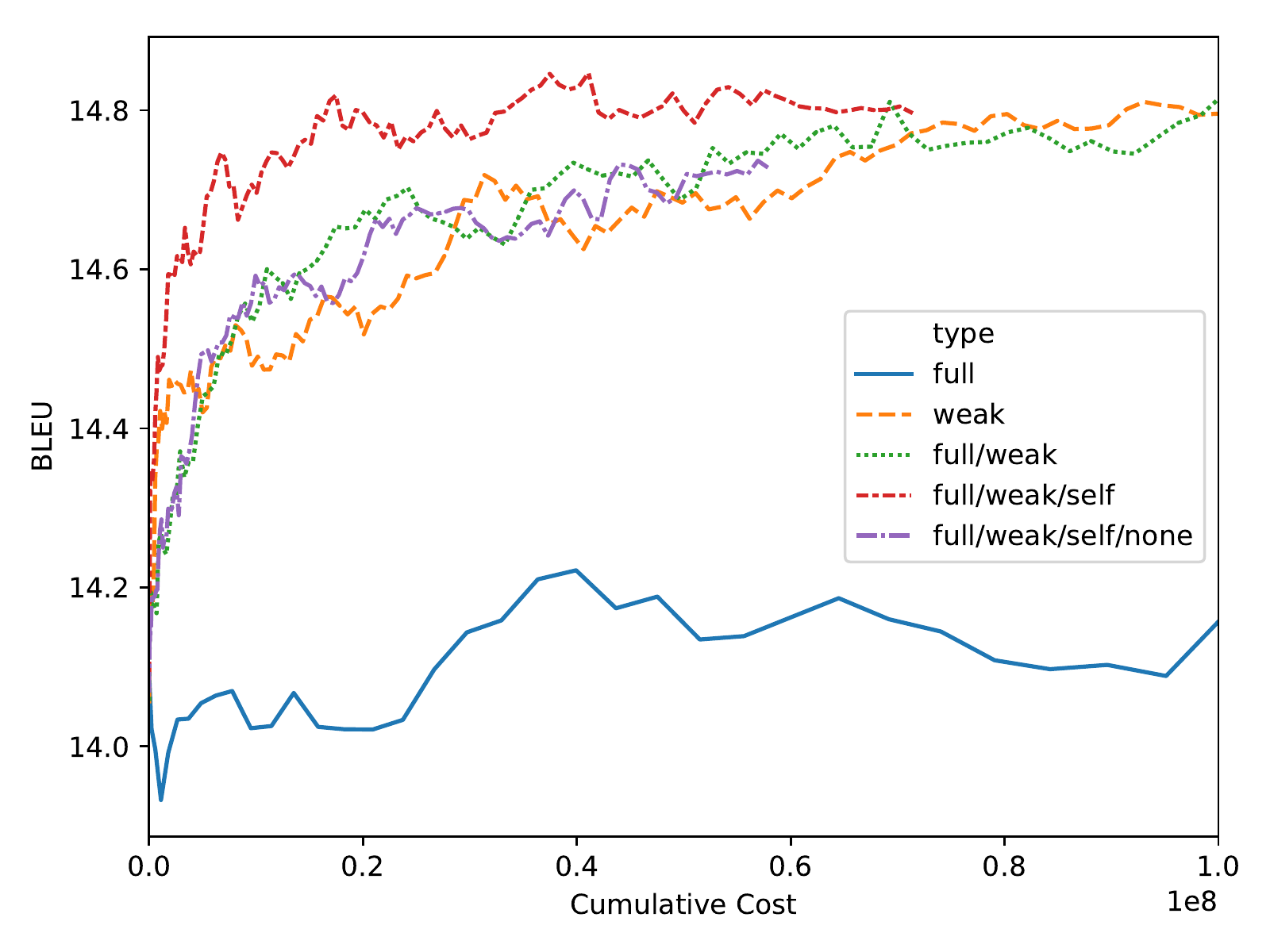}
    \caption{Domain transfer of regulators trained on IWSLT to the Books domain in comparison to full and weak feedback only.}
    \label{fig:books_cost}
\end{figure}
After training on IWSLT, we evaluate the regulators on the Books domain: 
Can they choose the best actions for an efficient learning progress without receiving feedback on the new domain?
We evaluate the best run of each regulator type (i.e., ${\phi}$ trained on IWSLT), with the Seq2Seq model reset to the WMT baseline. The regulator is not further adapted to the Books domain, but decides on the feedback types for training the Seq2Seq model for a single epoch on the Books data. Figure~\ref{fig:books_cost} visualizes the regulated training process of the Seq2Seq model. As before, \emph{Reg3} performs best, outperforming weak, full and self-supervision (reaching 14.75 BLEU, not depicted since zero cost). Learning from full feedback improves much later in training and reaches 14.53 BLEU.\footnote{With multiple epochs it would improve further, but we avoid showing the human the same inputs multiple times.} One explanation is that the reference translations in the Books corpus are less literal than the ones for IWSLT, such that a weak feedback signal allows the learner to learn more efficiently than from full corrections. Appendix~\ref{app:books-bleu} reports the results for offline evaluation on the trained Seq2Seq models on the Books test set.

\paragraph{Comparison to Active Learning.}
\begin{figure}[t]
    \centering
   \includegraphics[width=\columnwidth]{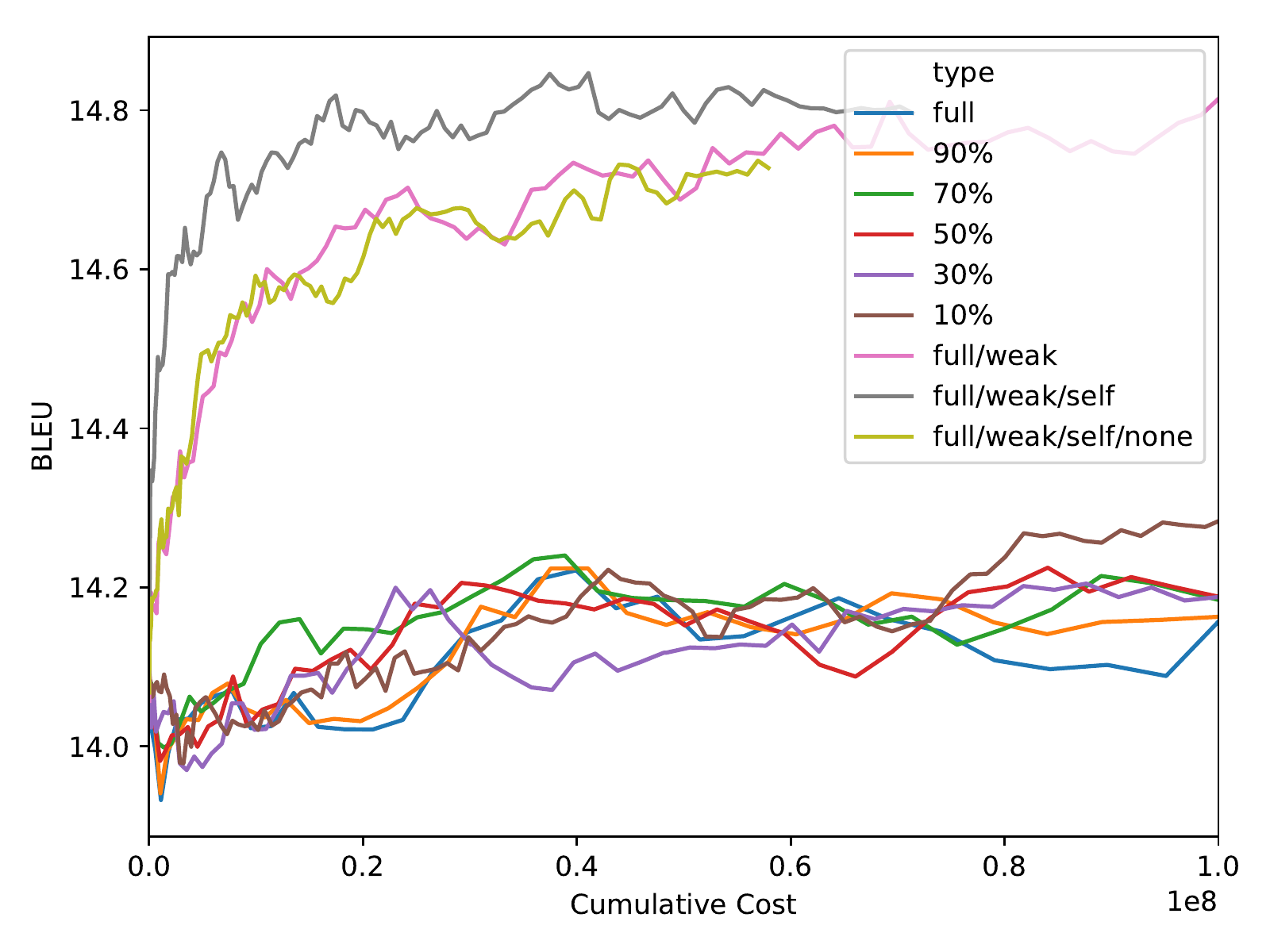}
    \caption{Learned self-regulation strategies in comparison to uncertainty-based active learning with a fixed percentage of full feedback on the Books domain.}
    \label{fig:al}
\end{figure}
A classic active learning strategy is to sample a subset of the input data for full labeling based on the uncertainty of the model predictions \cite{SettlesCraven:08}. The size of this subset, i.e. the amount of human labeling effort, has to be known and determined before learning. Figure~\ref{fig:al} compares the self-regulators on the Books domain with models that learn from a fixed ratio of fully-labeled instances in every batch. These are chosen according to the model's uncertainty, here measured by the average token entropy of the model's best-scoring beam search hypothesis. The regulated models with a mix of feedback types clearly outperform the active learning strategies, both in terms of cost-efficient learning (Figure~\ref{fig:al}) as well as in overall quality (See Figure~\ref{fig:al_time} in Appendix~\ref{app:al}). 
We conclude that mixing feedback types, especially in the case where full feedback is less reliable, offers large improvements over standard stream-based active learning strategies.

\subsection{Prospects for Field Studies}
Our experiments were designed as a pilot study to test the possibilities of self-regulated learning in simulation. In order to advance to field studies where human users interact with Seq2Seq models, several design choices have to be adapted with caution.
Firstly, we simulate both feedback cost and quality improvement by measuring distances to static reference outputs. The experimental design in a field study has to account for a variation of feedback strength, feedback cost, and performance assessments, across time, across sentences, and across human users \cite{SettlesETAL:08}. One desideratum for field studies is thus to analyze this variation by analyzing the experimental results in a mixed effects model that accounts for variability across sentences, users, and annotation sessions \cite{BaayenETAL:08,KarimovaETAL:18}. Secondly, our simulation of costs considers only the effort of the human teacher, not the machine learner. The strong preference for the cheapest feedback option might be a result of overestimating the cost of human post-editing and underestimating the cost of self-training. Thus, a model for field studies where data is limited might greatly benefit from learned estimates of feedback cost and quality improvement \cite{KreutzerETALacl:18}. 

\section{Conclusion}
We proposed a cost-aware algorithm for interactive sequence-to-sequence learning, with a self-regulation module at its core that learns which type of feedback to query from a human teacher. The empirical study on interactive NMT with simulated human feedback showed that this self-regulated model finds more cost-efficient solutions than models learning from a single feedback type and uncertainty-based active learning models, also under domain shift.
While this setup abstracts away from certain confounding variables to be expected in real-life interactive machine learning, it should be seen as a pilot experiment focused on our central research questions under an exact and noise-free computation of feedback cost and performance gain. The proposed framework can naturally be expanded to integrate more feedback modes suitable for the interaction with humans, e.g., pairwise comparisons or output rankings. 
Future research directions will involve the development of reinforcement learning model with multi-dimensional rewards, and modeling explicit credit assignment for improving the capabilities of the regulator to make context-sensitive decisions in mini-batch learning.

\section*{Acknowledgements}
We would like to thank the anonymous reviewers for their valuable feedback. The research reported in this paper was supported in part by the German research foundation (DFG) under grant RI-2221/4-1.

\bibliography{acl2019}

\begin{thebibliography}{49}
\expandafter\ifx\csname natexlab\endcsname\relax\def\natexlab#1{#1}\fi

\bibitem[{Baayen et~al.(2008)Baayen, Davidson, and Bates}]{BaayenETAL:08}
R~Harald Baayen, Douglas~J Davidson, and Douglas~M Bates. 2008.
\newblock Mixed-effects modeling with crossed random effects for subjects and
  items.
\newblock \emph{Journal of memory and language}, 59(4):390--412.

\bibitem[{Bahdanau et~al.(2015)Bahdanau, Cho, and Bengio}]{BahdanauETAL:15}
Dzmitry Bahdanau, Kyunghyun Cho, and Yoshua Bengio. 2015.
\newblock \href {https://arxiv.org/abs/1409.0473} {{Neural Machine Translation
  by Jointly Learning to Align and Translate}}.
\newblock In \emph{{International Conference on Learning Representations
  ({ICLR}})}, San Diego, California, USA.

\bibitem[{Barrachina et~al.(2009)Barrachina, Bender, Casacuberta, Civera,
  Cubel, Khadivi, Lagarda, Ney, Tom{\'a}s, Vidal, and
  Vilar}]{barrachinaETAL:09}
Sergio Barrachina, Oliver Bender, Francisco Casacuberta, Jorge Civera, Elsa
  Cubel, Shahram Khadivi, Antonio Lagarda, Hermann Ney, Jes{\'u}s Tom{\'a}s,
  Enrique Vidal, and Juan-Miguel Vilar. 2009.
\newblock \href {https://doi.org/10.1162/coli.2008.07-055-R2-06-29}
  {Statistical approaches to computer-assisted translation}.
\newblock \emph{Computational Linguistics}, 35(1).

\bibitem[{Bromley et~al.(1994)Bromley, Guyon, LeCun, S{\"a}ckinger, and
  Shah}]{bromley1994signature}
Jane Bromley, Isabelle Guyon, Yann LeCun, Eduard S{\"a}ckinger, and Roopak
  Shah. 1994.
\newblock Signature verification using a "siamese" time delay neural network.
\newblock In \emph{Advances in Neural Information Processing Systems
  {(NeurIPS)}}, Denver, {CO, USA}.

\bibitem[{Celemin et~al.(2019)Celemin, Ruiz-del Solar, and Kober}]{Celemin2019}
Carlos Celemin, Javier Ruiz-del Solar, and Jens Kober. 2019.
\newblock \href {https://doi.org/10.1007/s10514-018-9786-6} {A fast hybrid
  reinforcement learning framework with human corrective feedback}.
\newblock \emph{Autonomous Robots}, 43(5):1173--1186.

\bibitem[{Chen et~al.(2018)Chen, Firat, Bapna, Johnson, Macherey, Foster,
  Jones, Schuster, Shazeer, Parmar, Vaswani, Uszkoreit, Kaiser, Chen, Wu, and
  Hughes}]{bestofbothworlds}
Mia~Xu Chen, Orhan Firat, Ankur Bapna, Melvin Johnson, Wolfgang Macherey,
  George Foster, Llion Jones, Mike Schuster, Noam Shazeer, Niki Parmar, Ashish
  Vaswani, Jakob Uszkoreit, Lukasz Kaiser, Zhifeng Chen, Yonghui Wu, and
  Macduff Hughes. 2018.
\newblock The best of both worlds: Combining recent advances in neural machine
  translation.
\newblock In \emph{Proceedings of the 56th Annual Meeting of the Association
  for Computational Linguistics {(ACL)}}, Melbourne, Australia.

\bibitem[{Cheng et~al.(2018)Cheng, Yan, Wagener, and Boots}]{ChengETAL:18}
Ching-An Cheng, Xinyan Yan, Nolan Wagener, and Byron Boots. 2018.
\newblock Fast policy learning through imitation and reinforcement.
\newblock In \emph{Proceedings of the Conference on Uncertainty in Artificial
  Intelligence {(UAI)}}, Monterey, {CA, USA}.

\bibitem[{Clark et~al.(2018)Clark, Luong, Manning, and Le}]{ClarkETAL:2018}
Kevin Clark, Minh-Thang Luong, Christopher~D. Manning, and Quoc Le. 2018.
\newblock \href {http://aclweb.org/anthology/D18-1217} {Semi-supervised
  sequence modeling with cross-view training}.
\newblock In \emph{Proceedings of the 2018 Conference on Empirical Methods in
  Natural Language Processing {(EMNLP)}}, Brussels, Belgium.

\bibitem[{Domingo et~al.(2017)Domingo, Peris, and Casacuberta}]{Domingo2017}
Miguel Domingo, {\'A}lvaro Peris, and Francisco Casacuberta. 2017.
\newblock \href {https://doi.org/10.1007/s10590-017-9213-3} {Segment-based
  interactive-predictive machine translation}.
\newblock \emph{Machine Translation}, 31(4):163--185.

\bibitem[{Fang et~al.(2017)Fang, Li, and Cohn}]{FangETAL:17}
Meng Fang, Yuan Li, and Trevor Cohn. 2017.
\newblock \href {https://doi.org/10.18653/v1/D17-1063} {Learning how to active
  learn: A deep reinforcement learning approach}.
\newblock In \emph{Proceedings of the 2017 Conference on Empirical Methods in
  Natural Language Processing {(EMNLP)}}, Copenhagen, Denmark.

\bibitem[{Gehring et~al.(2017)Gehring, Auli, Grangier, Yarats, and
  Dauphin}]{gehring2017convolutional}
Jonas Gehring, Michael Auli, David Grangier, Denis Yarats, and Yann~N Dauphin.
  2017.
\newblock Convolutional sequence to sequence learning.
\newblock In \emph{International Conference on Machine Learning {(ICML)}},
  Vancouver, Canada.

\bibitem[{Hattie and Donoghue(2016)}]{HattieDonoghue:16}
John Hattie and Gregory~M. Donoghue. 2016.
\newblock Learning strategies: a synthesis and conceptual model.
\newblock \emph{NPJ Science of Learning}, 1:16013--16013.

\bibitem[{Hattie and Timperley(2007)}]{HattieTimperley:07}
John Hattie and Helen Timperley. 2007.
\newblock \href {https://doi.org/10.1038/npjscilearn.2016.13} {The power of
  feedback}.
\newblock \emph{American Educational Research Association}, 77(1):81--112.

\bibitem[{Hieber et~al.(2017)Hieber, Domhan, Denkowski, Vilar, Sokolov,
  Clifton, and Post}]{sockeye17}
Felix Hieber, Tobias Domhan, Michael Denkowski, David Vilar, Artem Sokolov, Ann
  Clifton, and Matt Post. 2017.
\newblock \href {http://arxiv.org/abs/1712.05690} {Sockeye: {A} toolkit for
  neural machine translation}.
\newblock \emph{CoRR}, abs/1712.05690.

\bibitem[{Hochreiter and Schmidhuber(1997)}]{hochreiter1997long}
Sepp Hochreiter and J{\"u}rgen Schmidhuber. 1997.
\newblock Long short-term memory.
\newblock \emph{Neural computation}, 9(8):1735--1780.

\bibitem[{Jaderberg et~al.(2017)Jaderberg, Mnih, Czarnecki, Schaul, Leibo,
  Silver, and Kavukcuoglu}]{JaderbergETAL:17}
Max Jaderberg, Volodymyr Mnih, Wojciech~Marian Czarnecki, Tom Schaul, Joel~Z.
  Leibo, David Silver, and Koray Kavukcuoglu. 2017.
\newblock Reinforcement learning with unsupervised auxiliary tasks.
\newblock In \emph{Proceedings of the International Conference on Learning
  Representations {(ICLR)}}, Toulon, France.

\bibitem[{Karimova et~al.(2018)Karimova, Simianer, and
  Riezler}]{KarimovaETAL:18}
Sariya Karimova, Patrick Simianer, and Stefan Riezler. 2018.
\newblock A user-study on online adaptation of neural machine translation to
  human post-edits.
\newblock \emph{Machine Translation}, 32(4):309--324.

\bibitem[{Kingma and Ba(2015)}]{KingmaBa:14}
Diederik Kingma and Jimmy Ba. 2015.
\newblock \href {https://arxiv.org/abs/1412.6980} {Adam: A method for
  stochastic optimization}.
\newblock In \emph{{International Conference on Learning Representations
  ({ICLR}})}, San Diego, CA, USA.

\bibitem[{{Kreutzer} et~al.(2019){Kreutzer}, {Bastings}, and
  {Riezler}}]{JoeyNMT}
Julia {Kreutzer}, Joost {Bastings}, and Stefan {Riezler}. 2019.
\newblock Joey {NMT}: A minimalist {NMT} toolkit for novices.
\newblock \emph{Proceedings of the 2019 Conference on Empirical Methods in
  Natural Language Processing and 9th International Joint Conference on Natural
  Language Processing: System Demonstrations {(EMNLP-IJCNLP)}}.

\bibitem[{Kreutzer et~al.(2018)Kreutzer, Uyheng, and
  Riezler}]{KreutzerETALacl:18}
Julia Kreutzer, Joshua Uyheng, and Stefan Riezler. 2018.
\newblock Reliability and learnability of human bandit feedback for
  sequence-to-sequence reinforcement learning.
\newblock In \emph{Proceedings of the 56th Annual Meeting of the Association
  for Computational Linguistics {(ACL)}}, Melbourne, Australia.

\bibitem[{Krueger et~al.(2016)Krueger, Leike, Evans, and
  Salvatier}]{KruegerETAL:16}
David Krueger, Jan Leike, Owain Evans, and John Salvatier. 2016.
\newblock Active reinforcement learning: Observing rewards at a cost.
\newblock In \emph{Proceeding of the 30th Conference on Neural Information
  Processing Systems {(NeurIPS)}}, Barcelona, Spain.

\bibitem[{Kumar et~al.(2019)Kumar, Foster, Cherry, and Krikun}]{KumarETAL:19}
Gaurav Kumar, George Foster, Colin Cherry, and Maxim Krikun. 2019.
\newblock Reinforcement learning based curriculum optimization for neural
  machine translation.
\newblock In \emph{Proceedings of the Annual Conference of the North American
  Chapter of the Association for Computational Linguistics {(NAACL)}},
  Minneapolis, {MN, USA}.

\bibitem[{Lam et~al.(2018)Lam, Kreutzer, and Riezler}]{LamETAL:18}
Tsz~Kin Lam, Julia Kreutzer, and Stefan Riezler. 2018.
\newblock A reinforcement learning approach to interactive-predictive neural
  machine translation.
\newblock In \emph{Proceedings of the 21st Annual Conference of the European
  Association for Machine Translation {(EAMT)}}, Alicante, Spain.

\bibitem[{Liu et~al.(2018)Liu, Buntine, and
  Haffari}]{liu-buntine-haffari:2018:K18-1}
Ming Liu, Wray Buntine, and Gholamreza Haffari. 2018.
\newblock \href {http://www.aclweb.org/anthology/K18-1033} {Learning to
  actively learn neural machine translation}.
\newblock In \emph{Proceedings of the 22nd Conference on Computational Natural
  Language Learning {(CoNLL)}}, Brussels, Belgium.

\bibitem[{Luong et~al.(2015)Luong, Pham, and
  Manning}]{luong-pham-manning:2015:EMNLP}
Thang Luong, Hieu Pham, and Christopher~D. Manning. 2015.
\newblock \href {http://aclweb.org/anthology/D15-1166} {Effective approaches to
  attention-based neural machine translation}.
\newblock In \emph{Proceedings of the 2015 Conference on Empirical Methods in
  Natural Language Processing {(EMNLP)}}, Lisbon, Portugal.

\bibitem[{MacGlashan et~al.(2017)MacGlashan, Ho, Loftin, Peng, Wang, Roberts,
  Taylor, and Littman}]{MacGlashanETAL:17}
James MacGlashan, Mark~K. Ho, Robert Loftin, Bei Peng, Guan Wang, David~L.
  Roberts, Matthew~E. Taylor, and Michael~L. Littman. 2017.
\newblock Interactive learning from policy-dependent human feedback.
\newblock In \emph{Proceedings of the 34th International Conference on Machine
  Learning {(ICML)}}, Sydney, Australia.

\bibitem[{Mitchell et~al.(2015)Mitchell, Cohen, Hruschka, Talukdar, Yang,
  Betteridge, Carlson, Dalvi, Gardner, Kisiel, Krishnamurthy, Lao, Mazaitis,
  Mohamed, Nakashole, Platanios, Ritter, Samadi, Settles, Wang, Wijaya, Gupta,
  Chen, Saparov, Greaves, and Welling}]{MitchellETAL:15}
T.~Mitchell, W.~Cohen, E.~Hruschka, P.~Talukdar, B.~Yang, J.~Betteridge,
  A.~Carlson, B.~Dalvi, M.~Gardner, B.~Kisiel, J.~Krishnamurthy, N.~Lao,
  K.~Mazaitis, T.~Mohamed, N.~Nakashole, E.~Platanios, A.~Ritter, M.~Samadi,
  B.~Settles, R.~Wang, D.~Wijaya, A.~Gupta, X.~Chen, A.~Saparov, M.~Greaves,
  and J.~Welling. 2015.
\newblock Never-ending learning.
\newblock In \emph{Proceedings of the 29th Conference on Artificial
  Intelligence {(AAAI)}}, Austin, {TX, USA}.

\bibitem[{Nigg(2017)}]{Nigg:17}
Joel~T. Nigg. 2017.
\newblock Annual research review: On the relations among self-regulation,
  self-control, executive functioning, effortful control, cognitive control,
  impulsivity, risk-taking, and inhibition for developmental psychopathology.
\newblock \emph{Journal of Child Psychology and Psychiatry}, 58(4):361--383.

\bibitem[{Panadero(2017)}]{Panadero:17}
Ernesto Panadero. 2017.
\newblock A review of self-regulated learning: Six models and four directions
  of research.
\newblock \emph{Frontiers in Psychology}, 8(422):1--28.

\bibitem[{Papineni et~al.(2002)Papineni, Roukos, Ward, and
  Zhu}]{papineni2002bleu}
Kishore Papineni, Salim Roukos, Todd Ward, and Wei-Jing Zhu. 2002.
\newblock Bleu: a method for automatic evaluation of machine translation.
\newblock In \emph{Proceedings of the 40th Annual Meeting on Association for
  Computational Linguistics (ACL)}, Philadelphia, {PA, USA}.

\bibitem[{Peris and Casacuberta(2018)}]{PerisCasacuberta:18}
{\'A}lvaro Peris and Francisco Casacuberta. 2018.
\newblock \href {http://www.aclweb.org/anthology/K18-1015} {Active learning for
  interactive neural machine translation of data streams}.
\newblock In \emph{Proceedings of the 22nd Conference on Computational Natural
  Language Learning (CONLL)}, Brussels, Belgium.

\bibitem[{Petrushkov et~al.(2018)Petrushkov, Khadivi, and
  Matusov}]{petrushkov-khadivi-matusov:2018:Short}
Pavel Petrushkov, Shahram Khadivi, and Evgeny Matusov. 2018.
\newblock \href {http://www.aclweb.org/anthology/P18-2052} {Learning from
  chunk-based feedback in neural machine translation}.
\newblock In \emph{Proceedings of the 56th Annual Meeting of the Association
  for Computational Linguistics {(ACL)}}, Melbourne, Australia.

\bibitem[{Post(2018)}]{sacrebleu}
Matt Post. 2018.
\newblock \href {http://aclweb.org/anthology/W18-6319} {A call for clarity in
  reporting {BLEU} scores}.
\newblock In \emph{Proceedings of the Third Conference on Machine Translation
  {(WMT)}}, Brussels, Belgium.

\bibitem[{Ranzato et~al.(2016)Ranzato, Chopra, Auli, and
  Zaremba}]{RanzatoETAL:16}
Marc'Aurelio Ranzato, Sumit Chopra, Michael Auli, and Wojciech Zaremba. 2016.
\newblock Sequence level training with recurrent neural networks.
\newblock In \emph{Proceedings of the International Conference on Learning
  Representation {(ICLR)}}, San Juan, Puerto Rico.

\bibitem[{Schmidhuber et~al.(1996)Schmidhuber, Zhao, and
  Wiering}]{SchmidhuberETAL:96}
J\"urgen Schmidhuber, Jieyu Zhao, and Marco Wiering. 1996.
\newblock Simple principles of metalaerning.
\newblock Technical Report 69 96, {IDSIA}, Lugano, Switzerland.

\bibitem[{Sennrich et~al.(2016)Sennrich, Haddow, and Birch}]{BPE}
Rico Sennrich, Barry Haddow, and Alexandra Birch. 2016.
\newblock Neural machine translation of rare words with subword units.
\newblock In \emph{Proceedings of the 54th Annual Meeting of the Association
  for Computational Linguistics {(ACL)}}, Berlin, Germany.

\bibitem[{Settles and Craven(2008)}]{SettlesCraven:08}
Burr Settles and Mark Craven. 2008.
\newblock An analysis of active learning strategies for sequence labeling
  tasks.
\newblock In \emph{Proceedings of the Conference on Empirical Methods in
  Natural Language Processing {(EMNLP)}}, Honolulu, Hawaii.

\bibitem[{Settles et~al.(2008)Settles, Craven, and Friedland}]{SettlesETAL:08}
Burr Settles, Mark Craven, and Lewis Friedland. 2008.
\newblock Active learning with real annotation costs.
\newblock In \emph{Proceedings of the {NeurIPS} Workshop on Cost-Sensitive
  Learning}, Vancouver, Canada.

\bibitem[{Smith et~al.(2018)Smith, Hoof, and Pineau}]{SmithETAL:18}
Matthew J.~A. Smith, Herke~Van Hoof, and Joelle Pineau. 2018.
\newblock An inference-based policy gradient method for learning options.
\newblock In \emph{Proceedings of the 35th International Conference on Machine
  Learning {(ICML)}}, Stockholm, Sweden.

\bibitem[{Snover et~al.(2006)Snover, Dorr, Schwartz, Micciulla, and
  Makhoul}]{snover2006study}
Matthew Snover, Bonnie Dorr, Richard Schwartz, Linnea Micciulla, and John
  Makhoul. 2006.
\newblock A study of translation edit rate with targeted human annotation.
\newblock In \emph{Proceedings of association for machine translation in the
  Americas (AMTA)}, volume 200.

\bibitem[{Sutskever et~al.(2014)Sutskever, Vinyals, and
  Le}]{sutskever2014sequence}
Ilya Sutskever, Oriol Vinyals, and Quoc~V Le. 2014.
\newblock Sequence to sequence learning with neural networks.
\newblock In \emph{Advances in neural information processing systems
  {(NeurIPS)}}, Montreal, Canada.

\bibitem[{Thrun and Pratt(1998)}]{ThrunPratt:98}
Sebastian Thrun and Lorien Pratt, editors. 1998.
\newblock \emph{Learning to Learn}.
\newblock Kluwer, Dortrecht, {MA, USA}.

\bibitem[{Tiedemann(2012)}]{TIEDEMANN12.463}
Jörg Tiedemann. 2012.
\newblock Parallel data, tools and interfaces in opus.
\newblock In \emph{Proceedings of the Eight International Conference on
  Language Resources and Evaluation (LREC)}, Istanbul, Turkey.

\bibitem[{Turchi et~al.(2017)Turchi, Negri, Farajian, and
  Federico}]{turchi2017continuous}
Marco Turchi, Matteo Negri, M~Amin Farajian, and Marcello Federico. 2017.
\newblock Continuous learning from human post-edits for neural machine
  translation.
\newblock \emph{The Prague Bulletin of Mathematical Linguistics},
  108(1):233--244.

\bibitem[{Vaswani et~al.(2017)Vaswani, Shazeer, Parmar, Uszkoreit, Jones,
  Gomez, Kaiser, and Polosukhin}]{vaswani2017attention}
Ashish Vaswani, Noam Shazeer, Niki Parmar, Jakob Uszkoreit, Llion Jones,
  Aidan~N Gomez, {\L}ukasz Kaiser, and Illia Polosukhin. 2017.
\newblock Attention is all you need.
\newblock In \emph{Advances in Neural Information Processing Systems
  {(NeurIPS)}}, Long Beach, {CA, USA}.

\bibitem[{Watkins(1989)}]{watkins1989learning}
Christopher Watkins. 1989.
\newblock Learning from delayed rewards.
\newblock \emph{PhD thesis, Cambridge University}.

\bibitem[{Williams(1992)}]{Williams:92}
Ronald~J. Williams. 1992.
\newblock Simple statistical gradient-following algorithms for connectionist
  reinforcement learning.
\newblock \emph{Machine Learning}, 8:229--256.

\bibitem[{Wun et~al.(2018)Wun, Tian, Xia, Fan, Qin, Lai, and Liu}]{WuETAL:18}
Lijun Wun, Fei Tian, Yingce Xia, Yang Fan, Tao Qin, Jianhuang Lai, and Tie-Yan
  Liu. 2018.
\newblock Learning to teach with dynamic loss functions.
\newblock In \emph{Proceeding of the 32nd Conference on Neural Information
  Processing System {(NeuRIPS)}}, Montreal, Canada.

\bibitem[{Zimmerman and Schunk(1989)}]{ZimmermanSchunk:89}
Barry~J. Zimmerman and Dale~H. Schunk, editors. 1989.
\newblock \href {https://doi.org/10.1007/978-1-4612-3618-4}
  {\emph{Self-Regulated Learning and Academic Achievement}}.
\newblock Springer, New York, NY, USA.

\end{thebibliography}
\bibliographystyle{acl_natbib}

\appendix

\newpage

\section{Appendices}
\label{sec:supplemental}

\subsection{Data}\label{app:data}
\begin{table}[h!]
\centering
\begin{tabular}{l|ccc}
\toprule
de$\leftrightarrow$en & \textbf{WMT}  & \textbf{IWSLT }& \textbf{Books} \\
\midrule
Train & 5,889,699 & 206,112 & 46,770\\
Dev & 2,169 &  2,385 & 2,000\\
Test & 3,004 & 1,138 & 2,000\\
\bottomrule
\end{tabular}
\caption{Number of sentences for parallel corpora used for pre-training (WMT), regulator training (IWSLT) and domain transfer evalution (Books).}
\label{tab:data}
\end{table}

The WMT data is obtained from the WMT 2017 shared task website\footnote{\url{http://www.statmt.org/wmt17/translation-task.html}} and pre-processed as described in \citet{sockeye17}.
The pre-processing pipeline is used for IWSLT and Books data as well. IWSLT2017 is obtained from the evaluation campaign website.\footnote{\url{https://sites.google.com/site/iwsltevaluation2017/}} For validation on WMT, we use the \texttt{newstest2015} data, for IWSLT \texttt{tst2014+tst2015}, for testing on WMT \texttt{newstest2017} and \texttt{tst2017} for IWSLT. Since there is no standard split for the Books corpus, we randomly select 2k sentences for validation and testing each. Table~\ref{tab:data} gives an overview of the size of the three resources.

\subsection{Online Evaluation on IWSLT}\label{app:iwslt}
\begin{figure*}[h]
   \begin{subfigure}{\columnwidth}
     \centering
   \includegraphics[width=\columnwidth]{rnn-full-regs-all.pdf}
  \caption{BLEU over cumulative costs.}
      \label{fig:bleu-cost-app}
   \end{subfigure}
           \hfill
    \begin{subfigure}{\columnwidth}
      \centering
   \includegraphics[width=\columnwidth]{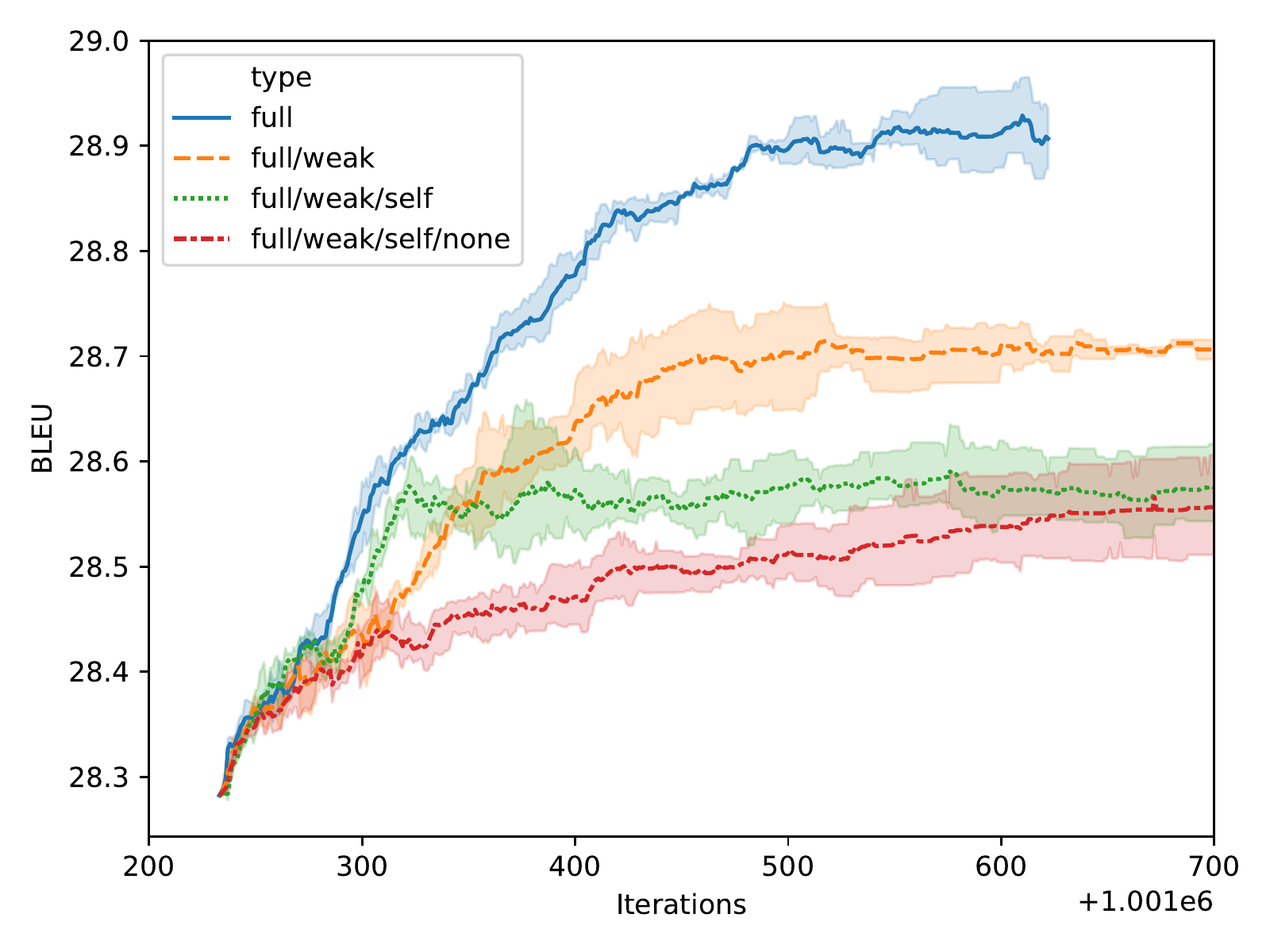}
    \caption{BLEU over time.}
 \label{fig:bleu-time-app}
   \end{subfigure}
    \caption{Regulation variants evaluated in terms of BLEU over time (a) and cumulative costs (b). Iteration counts start from the iteration count of the baseline model. One iteration on IWSLT equals training on one mini-batch of 32 instances. The BLEU score is computed on the tokenized validation set with greedy decoding. In (b) the lines correspond to the means over three runs, the shaded area depicts the estimated 95\% confidence interval.}
    \label{fig:reg_variants_full_app}
\end{figure*}

Figure~\ref{fig:reg_variants_full_app} displays the development of BLEU over costs and time.

\subsection{Offline Evaluation on IWSLT}\label{app:offline}

\begin{table}[h]
    \centering
   \resizebox{\columnwidth}{!}{
    \begin{tabular}{c|cc|cc}
        \toprule
         \textbf{Model }& \multicolumn{2}{c|}{\textbf{IWSLT dev}} & \multicolumn{2}{c}{\textbf{IWSLT test}}\\
         & \textbf{BLEU}$\uparrow$ & \textbf{Cost}$\downarrow$ & \textbf{BLEU}$\uparrow$  & \textbf{TER}$\downarrow$  \\
         \midrule
         \emph{Baseline} & 28.28 & - & 24.84 & 62.42 \\
         \midrule
         \emph{Full} & 28.93$_{\pm 0.02}$  &417k & 25.60$_{\pm 0.02}$ & 61.86$_{\pm 0.03}$  \\ 
        \emph{Weak} & 28.65$_{\pm 0.01}$ & 32k & 25.10$_{\pm 0.09}$ & 62.12$_{\pm 0.12}$ \\ 
        \emph{Self} & 28.58$_{\pm 0.02}$ & - & 25.33$_{\pm 0.06}$ & 61.96$_{\pm 0.05}$ \\ 
         \midrule
         \emph{Reg4} & 28.57$_{\pm 0.04}$ & 68k & 25.23$_{\pm 0.05}$ & 62.02$_{\pm 0.12}$  \\ 
          \emph{Reg3} & 28.61$_{\pm 0.03}$ & 18k & 25.23$_{\pm 0.09}$ & 62.07$_{\pm 0.06}$  \\
        \emph{Reg2} & 28.66$_{\pm 0.06}$ & 88k & 25.27$_{\pm 0.09}$  & 61.91$_{\pm 0.06}$  \\
         \bottomrule
    \end{tabular}
    }
    \caption{Evaluation of models at early stopping points. Results for three random seeds on IWSLT are averaged, reporting the standard deviation in the subscript. The translation of the dev set is obtained by greedy decoding (as during validation) and of the test set with beam search of width five.
    The costs are measured in character edits and clicks, as described in Section~\ref{sec:experiments}.}
    \label{tab:results}
\end{table}

Table~\ref{tab:results} reports the offline held-out set evaluations for the early stopping points selected on the dev set for all feedback modes. All models notably improve over the baseline, only using full feedback leads to the overall best model on IWSLT (+0.6 BLEU / -0.6 TER), but costs a massive amounts of edits (417k characters). Self-regulating models still achieve improvements of 0.4--0.5 BLEU/TER with costs reduced up to a factor of 23. The reduction in cost is enabled by the use of cheaper feedback, here markings and self-supervision, which in isolation are successful as well. Self-supervision works surprisingly well, which makes it attractive for cheap but effective unsupervised domain adaptation. It has to be noted that both weak and self-supervision worked only well when targets were pre-computed with the baseline model and held fixed during training. We suspect that the strong reward signal ($f_t=1$) for non-reference outputs leads otherwise to undesired local overfitting effects that a learner with online-generated targets cannot recover from.

\subsection{Domain Transfer}\label{app:books-bleu}
\begin{table}[h]
    \centering
   \resizebox{0.7\columnwidth}{!}{
    \begin{tabular}{c|ccc}
        \toprule
         \textbf{Model} & \multicolumn{3}{c}{\textbf{Books test}}  \\
          & \textbf{BLEU}$\uparrow$  & \textbf{TER}$\downarrow$ & \textbf{Cost}$\downarrow$ \\
         \midrule
         \emph{Baseline} & 14.19  & 79.81 & - \\
         \midrule
         \emph{Full} & 14.87  & 79.12 & 1B \\ 
        \emph{Weak} & 14.74  & 78.14 & 93M \\ 
        \emph{Self} & 14.73 & 78.86 & - \\ 
         \midrule
         \emph{Reg4} & 14.80 & 79.02 & 57M \\
         \emph{Reg3} & 14.80 & 78.70 & 41M \\
        \emph{Reg2} & 15.00 & 78.21 & 142M \\ 
         \bottomrule
    \end{tabular}
    }
    \caption{Evaluation of models at early stopping points on the Books test set (beam search with width five).}
    \label{tab:results_books}
\end{table}
Table \ref{tab:results_books} reports results for test set evaluation on the Books domain of the best model from the IWSLT domain each. The baseline was trained on WMT parallel data without any regulation. The regulator was trained on IWSLT and evaluated on Books, the Seq2Seq model is further trained for one epoch on Books. The costs are measured in character edits and clicks. The best result in terms of BLEU and TER is achieved by the \emph{Reg2} model, even outperforming the model with full feedback. As observed for the IWSLT domain (cf. Section~\ref{sec:results}), self-training is very effective, but is outperformed by the \emph{Reg2} model and roughly on par with the \emph{Reg3} model.

\subsection{Active Learning on Books}\label{app:al}
\begin{figure}[h]
    \centering
    \includegraphics[width=\columnwidth]{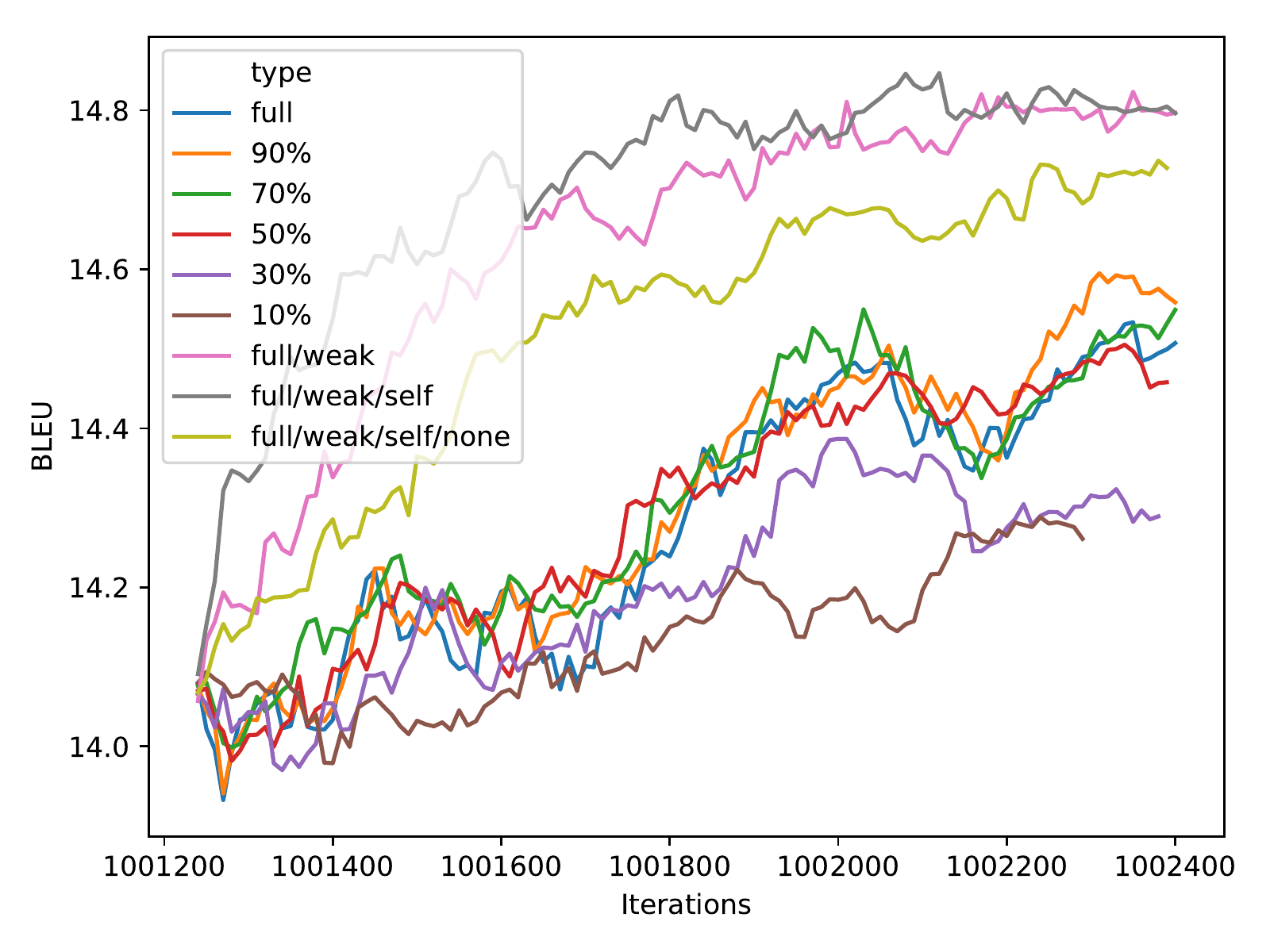}
    \caption{Development of validation BLEU over time for learned regulation strategies in comparison to active learning with a fixed percentage $\gamma$ of full feedback. Counting of iterations starts at the previous iteration count of the baseline model.}
    \label{fig:al_time}
\end{figure}
Figure~\ref{fig:al_time} shows the development of BLEU over time for the regulators and active learning strategies with a fixed ratio of full feedback per batch ($\gamma \in [10,30,50,70,90]$). The decision whether to label an instance in a batch is made based on the average token entropy of the model's current hypothesis. Using only 50\% of the fully-supervised labels achieves the same quality as 100\% using this uncertainty-based active learning sampling strategy. However, the regulated models reach a higher quality not only at a lower cost (see Figure~\ref{fig:al}), but also reach an overall higher quality. 

\subsection{Regulation Strategies on IWSLT}\label{app:iwslt_reg}
\begin{figure}[h]
    \centering
       \includegraphics[width=0.8\columnwidth]{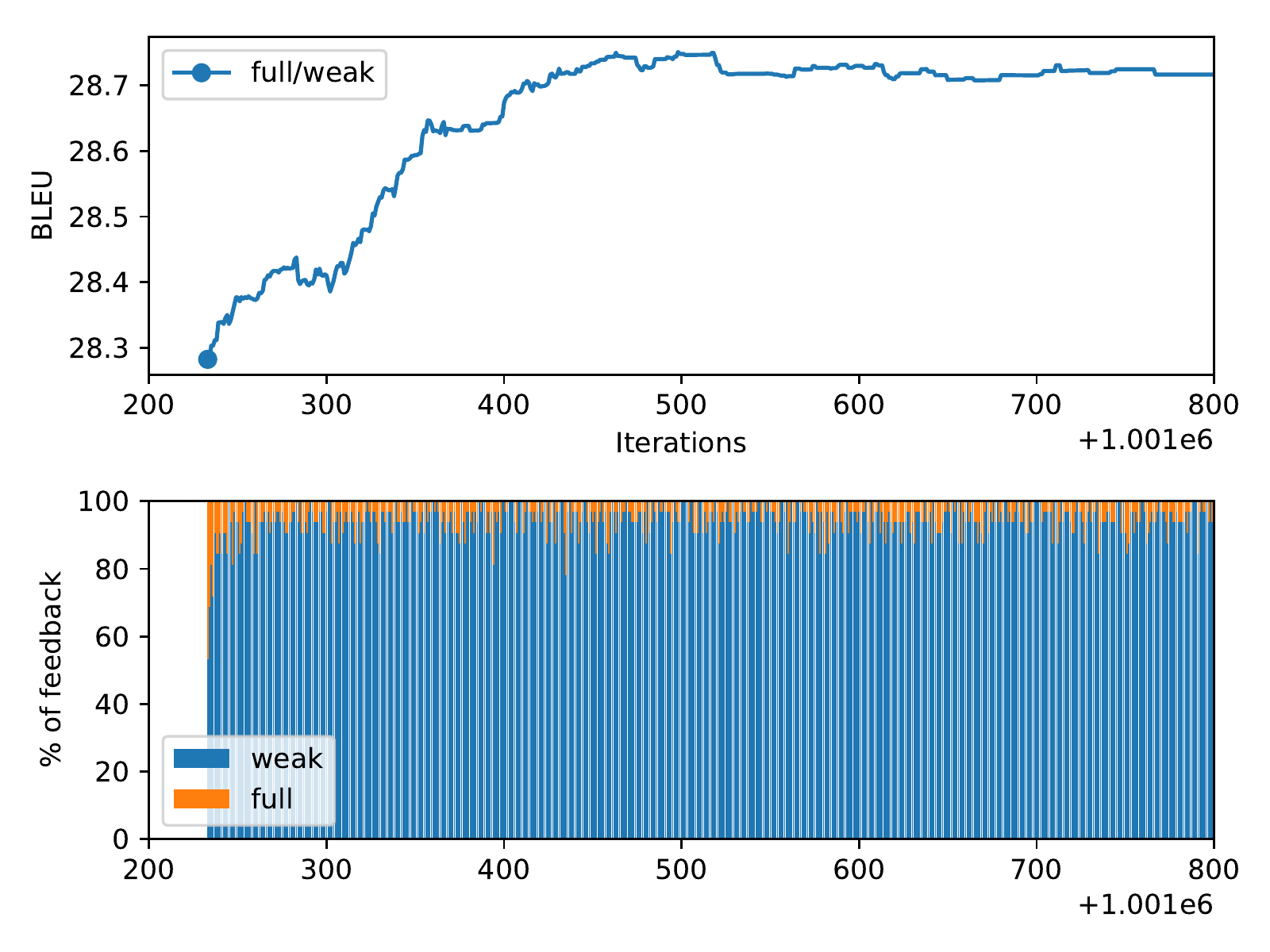}
    \caption{Feedback chosen by \emph{Reg2} on IWSLT.}
    \label{fig:reg2_iwslt}
\end{figure}
\begin{figure}[h]
    \centering
   \includegraphics[width=0.8\columnwidth]{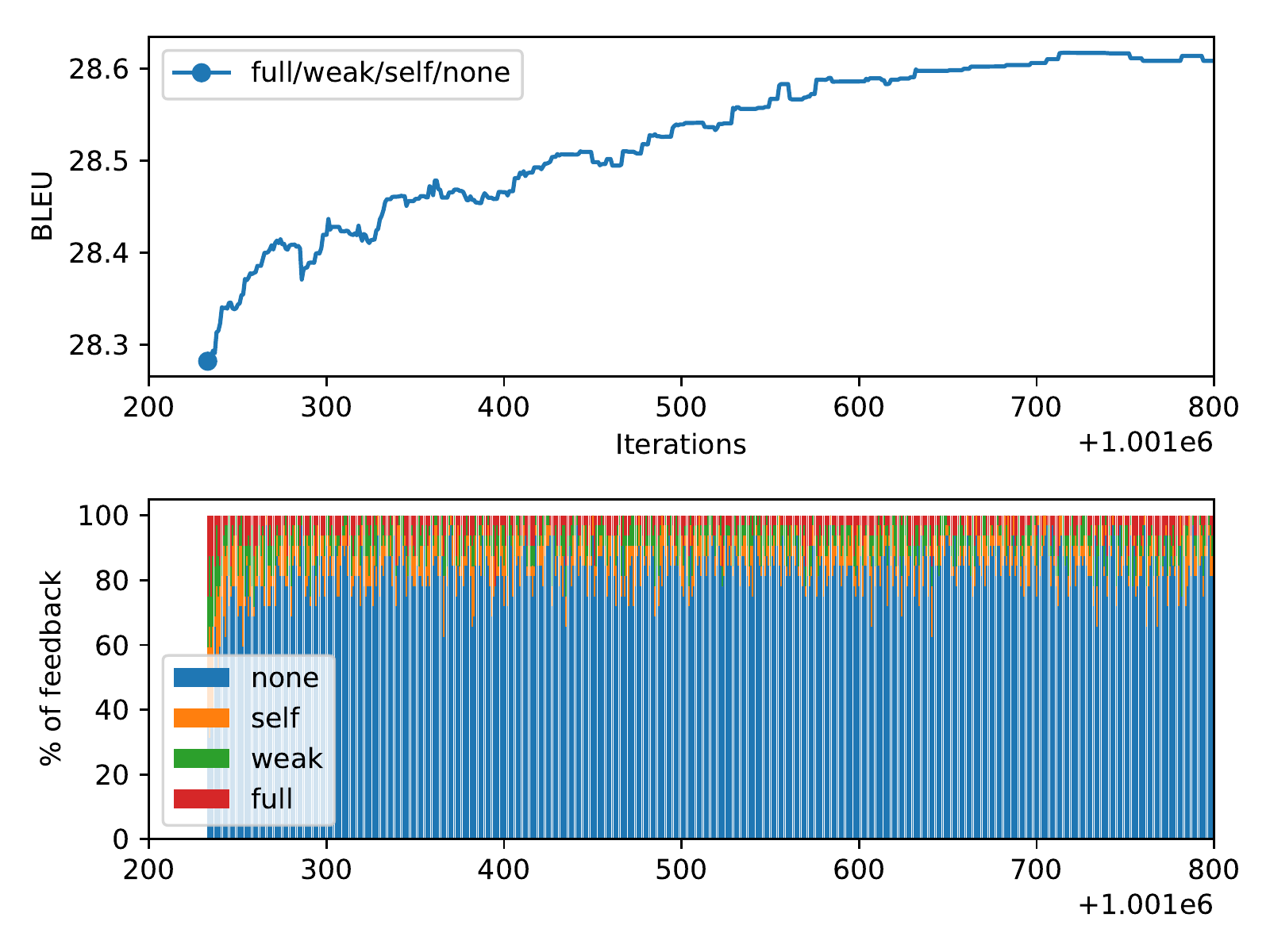}
    \caption{Feedback chosen by \emph{Reg4} on IWSLT.}
    \label{fig:reg4_iwslt}
\end{figure}
Figures~\ref{fig:reg2_iwslt} and \ref{fig:reg4_iwslt} show the ratio of feedback types for self-regulation during training with \emph{Reg2} and \emph{Reg4} respectively.

\end{document}